\documentclass[10pt,twocolumn,letterpaper]{article}

\usepackage{cvpr}              

\usepackage[dvipsnames]{xcolor}

\definecolor{cvprblue}{rgb}{0.21,0.49,0.74}
\usepackage[pagebackref,breaklinks,colorlinks,citecolor=cvprblue]{hyperref}
\usepackage{graphicx}
\usepackage{pifont}

\usepackage{caption} 
\usepackage{booktabs} 
\usepackage{xcolor} 

\definecolor{codegreen}{rgb}{0,0.4,0}
\definecolor{codegray}{rgb}{1.0,0.5,0.5}
\definecolor{codepurple}{rgb}{0.58,0,0}
\definecolor{tealblue}{rgb}{0,0.5,0.5}
\definecolor{codebackcolour}{rgb}{0.95,0.95,0.92}
\definecolor{darkgreen}{RGB}{0,127,0}
\definecolor{darkred}{RGB}{200,0,0}
\definecolor{orange}{rgb}{1,0.5,0}
\def\greencheckmark{\textcolor{darkgreen}{\checkmark}}
\def\redxmark{\textcolor{darkred}{\ding{55}}}  

\definecolor{grey50}{rgb}{0.5,0.5,0.5}

\newcommand{\genca}{\textcolor{black}{GenCA }}

\newcommand{\decblock }{\mathcal{D}}
\newcommand{\zgeo}{z_{\scriptsize \mbox{geo}}}
\newcommand{\ztex}{z_{\scriptsize \mbox{tex}}}
\newcommand{\zexp}{z_{\scriptsize \mbox{exp}}}
\newcommand{\Eexp}{f_{\scriptsize \mbox{exp}}}
\newcommand{\Egeo}{f_{\scriptsize \mbox{geo}}}
\newcommand{\Dgeo}{g_{\scriptsize \mbox{geo}}}
\newcommand{\Etex}{f_{\scriptsize \mbox{tex}}}
\newcommand{\Dtex}{g_{\scriptsize \mbox{tex}}}
\newcommand{\Dexp}{g_{\scriptsize \mbox{exp}}}
\newcommand{\prompt}{\mathcal{P}}
\newcommand{\renderer}{\mathcal{R}}
\newcommand{\hypernet}{\mathcal{H}}
\newcommand{\Gneu}{\mathcal{G}_{\scriptsize \mbox{neu}}}
\newcommand{\Tneu}{\mathcal{T}_{\scriptsize \mbox{neu}}}
\newcommand{\Gexp}{\mathcal{G}_{\scriptsize \mbox{exp}}}
\newcommand{\Texp}{\mathcal{T}_{\scriptsize \mbox{exp}}}
\newcommand{\Irecon}{\hat{I}}
\newcommand{\campose}{\mathcal{C}}
\newcommand{\registration}{\mathbf{\mathfrak{R}}}

\newcommand\blfootnote[1]{
    \begingroup
    \renewcommand\thefootnote{}\footnote{#1}
    \addtocounter{footnote}{-1}
    \endgroup
}

\def\paperID{131} 
\def\confName{3DV\xspace}
\def\confYear{2025\xspace}

\title{GenCA: A Text-conditioned Generative Model for Realistic and Drivable Codec Avatars}

\author{
  Keqiang Sun$^1$,
  Amin Jourabloo$^2$,
  Riddhish Bhalodia$^2$,
  Moustafa Meshry$^2$,
  Yu Rong$^2$,
  \and
  Zhengyu Yang$^2$,
  Thu Nguyen-Phuoc$^2$,
  Christian Haene$^2$,
  Jiu Xu$^2$,
  Sam Johnson$^2$,
  \and
  Hongsheng Li$^1$,
  Sofien Bouaziz$^2$\\ \\
  $^1$ Chinese University of Hong Kong, $^2$ Meta Reality Labs 
}

\begin{document}

\twocolumn[
\maketitle
\vspace{-3em}
\begin{center} 
  \includegraphics[trim = 0 0 0 0, width=1\linewidth]{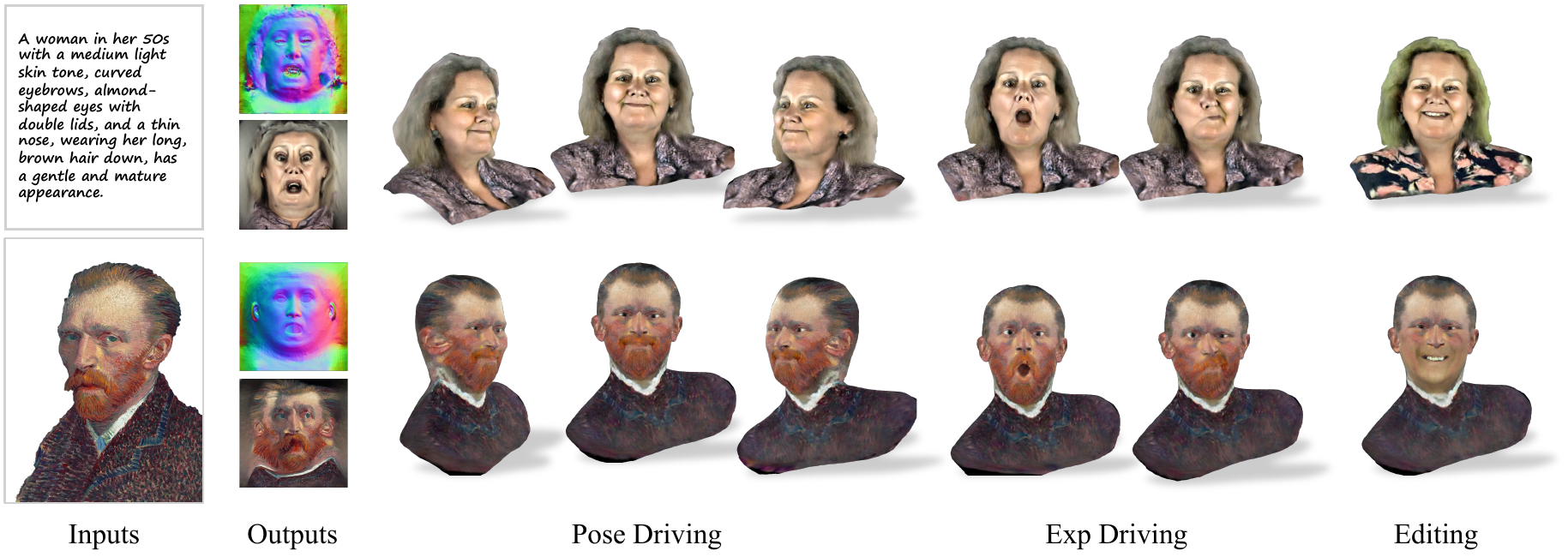}
\captionof{figure}{\textbf{Generative Codec Avatars.} Given a sentence describing the attributes of a face, our method generates a Codec Avatar, which can be driven by realistic expressions (top). \genca has many downstream applications such as avatar reconstruction from a single in-the-wild image (bottom). Additionally, it allows for editing features, such as changing hair color to green (top) or removing facial hair (bottom).
  }
\label{fig:teaser}
\end{center}
\medbreak
]

\begin{abstract}

\blfootnote{This work was performed during the first author’s internship at
Meta, Burlingame, CA.}
    Photo-realistic and controllable 3D avatars are crucial for various applications such as virtual and mixed reality (VR/MR), telepresence, gaming, and film production. 
    Traditional methods for avatar creation often involve time-consuming scanning and reconstruction processes for each avatar, which limits their scalability.
    Furthermore, these methods do not offer the flexibility to sample new identities or modify existing ones.
    On the other hand, by learning a strong prior from data, generative models provide a promising alternative to traditional reconstruction methods, easing the time constraints for both data capture and processing.
    Additionally, generative methods enable downstream applications beyond reconstruction, such as editing and stylization.
    Nonetheless, the research on generative 3D avatars is still in its infancy, and therefore current methods still have limitations such as creating static avatars, lacking photo-realism, having incomplete facial details, or having limited drivability.
    To address this, we propose a text-conditioned generative model that can generate photo-realistic facial avatars of diverse identities, with more complete details like hair, eyes and mouth interior, and which can be driven through a powerful non-parametric latent expression space.
    Specifically, we integrate the generative and editing capabilities of latent diffusion models with a strong prior model for avatar expression driving.
    Our model can generate and control high-fidelity avatars, even those out-of-distribution. We also highlight its potential for downstream applications, including avatar editing and single-shot avatar reconstruction.

  
\end{abstract}
\section{Introduction}
\label{sec:intro}





Generating high-quality human face models has numerous applications in the gaming and film industries.
Recently, social telepresence applications in virtual reality (VR) and mixed reality (MR) have created new demands for highly-accurate and authentic avatars that can be controlled by users' input expressions.
These avatars play a vital role in improving user experience and immersion in VR and MR, making their development an area of significant interest. 

Current methods for creating 3D avatars can be categorized into reconstruction-based and generative-based approaches.
Reconstruction-based methods, such as the \emph{Codec Avatar} family of works~\cite{ma2021pixelca,saito2023relightable}, recover highly photo-realistic 3D avatars, but mostly rely on extensive multi-view captures of real humans.
Additionally, they require a lengthy reconstruction process.
Recently, \cite{cao2022authentic} has reduced the need for an extensive capture by training a Universal Prior Model (UPM) using high-quality multi-view captures, and subsequently fine-tuning this learned prior with a person-specific phone scan.
A state-of-the-art instant avatar method~\cite{zielonka2023insta} made a significant stride by further relaxing the capture requirement to a monocular HD video, and reduced the avatar reconstruction time down to 10 minutes.
%
However, these methods only reconstruct a 3D representation that replicates the identity and appearance of a given human performance capture,
but support neither single-/few-shot reconstructions, nor editing capability, and cannot generate fictional avatars (\emph{e.g.,} for the gaming and movie industries). 

On the other hand, generative models, especially conditional diffusion models, have demonstrated remarkable capabilities in generating high-quality photo-realistic images from various conditional signals. 
%
These 2D image generative models can be used to generate 3D avatars~\cite{describe3d, tada, dreamface} and have shown promising results for generating and \emph{editing} high-quality avatars from text descriptions.
However, the generated avatars are not photo-realistic and have limited completeness for areas such as eyes, mouth interior, hair, and wearable accessories.
Moreover, to create a single avatar, these methods still rely on a lengthy optimization or distillation process even for state-of-the-art methods, such as DreamFace~\cite{dreamface}.
Other 3D generative models~\cite{rodin, Chan_2021_CVPR, Niemeyer2020GIRAFFE, eg3d} recover an implicit 3D representation that can be rendered from input camera views into photo-realistic images.
While these methods generate photo-realistic face avatars with good completeness (\emph{e.g.,} hair, teeth, glasses and other accessories), the generated avatars are static and cannot be driven by users' expressions.
%
%
%
Therefore, in this work, we combine the authenticity and drivability of Codec Avatars~\cite{cao2022authentic,lombardi2021mixture}, the generalization and completeness of 3D generative models~\cite{rodin,eg3d}, and the intuitive text-based editing capability of 2D vision-language generative models~\cite{dreamface} (see \Cref{tab:sotaconcept}).

We propose Generative Codec Avatars (\genca), a two-stage framework for generating drivable 3D avatars using only text descriptions. In the first stage, we introduce the Codec Avatar Auto-Encoder (CAAE), which learns geometry and texture latent spaces from a dataset of 3D human captures. These latent spaces model the identity distribution of avatars and are combined with an expression latent space from a Universal Prior Model (UPM)~\cite{cao2022authentic} to enable expression-driven, high-quality rendering of the generated identities. In the second stage, we present the Identity Generation Model. Here, the Geometry Generation module learns to generate the neutral geometry code based on the input text prompt, while the Geometry Conditioned Texture Generation Module learns to generate the neutral texture conditioned on both the geometry and the text.
%
The generated drivable avatars capture a far more complete representation of human heads (Fig.~\ref{fig:teaser}-top) compared to prior state-of-the-art \emph{generative drivable avatars}~\cite{dreamface, tada, describe3d}.
Additionally, our method significantly improves the driving capabilities of the generated avatars, including the ability to control areas like the eyes and tongue.
Those areas are neither represented nor controlled in previous methods \emph{generative drivable avatars}, which rely on parametric face models~\cite{blanz1999morphable}.
%
To demonstrate the effectiveness of our learned prior, we adapt an inversion process from~\cite{xia2022gan, gal2023an} to enable drivable avatar reconstruction from a single in-the-wild image (Fig.~\ref{fig:teaser}-bottom).
We further demonstrate avatar editing results, beyond the training data, for both reconstructed and generated avatars (Fig.~\ref{fig:teaser}-last column).
%
%
In summary, our contributions are:
%

\begin{itemize}
    \item We present \genca, the first text-conditioned generative model for photo-realistic, editable, and free-form drivable 3D avatar generation.
    \item We devise a Codec Avatar Auto-encoder to map facial images into the latent space, and the Identity Generation Model for the Codec Avatar generation.
    \item We showcase a variety of downstream applications enabled by \genca model, including 3D avatar reconstruction from a single image, avatar editing and inpainting. %

\end{itemize}    

\begin{table}
    \centering
    \caption{Comparison between our proposed method (GenCA) and state-of-the-art avatar creation methods.}
    \resizebox{1.00\linewidth}{!}{
    \begin{tabular}{lccccc}
        \toprule
          & Generative & Photo-real & Completeness & Drivablity & Editability \\
         \midrule
         PanoHead~\cite{an2023panohead} & \greencheckmark & \greencheckmark & \greencheckmark & \redxmark & \greencheckmark \\
         RODIN~\cite{rodin} & \greencheckmark & \redxmark & \greencheckmark & \redxmark & \greencheckmark \\
         ICA~\cite{ICA} & \redxmark & \greencheckmark & \greencheckmark & \greencheckmark & \redxmark \\
         INSTA~\cite{zielonka2023insta} & \redxmark & \greencheckmark & \greencheckmark & \greencheckmark & \redxmark \\
         Describ3D~\cite{describe3d} & \greencheckmark & \redxmark & \redxmark & \greencheckmark & \greencheckmark \\
         TADA~\cite{tada} & \redxmark & \redxmark & \redxmark & \greencheckmark & \greencheckmark \\
         DreamFace~\cite{dreamface} & \redxmark & \greencheckmark & \redxmark & \greencheckmark & \greencheckmark \\
         \textbf{\genca (ours)} & \greencheckmark & \greencheckmark & \greencheckmark & \greencheckmark & \greencheckmark \\
         \bottomrule
    \end{tabular}
    }
    \label{tab:sotaconcept}
\end{table}

\section{Related Works}
\label{sec:literature}
Reconstruction or generating photo-realistic 3D face avatars is a well-studied problem in computer graphics and computer vision.
Existing solutions often make trade-offs along different axes, such as avatar quality, model completeness, reconstruction/generation cost, driveability, editability, and generative capability.
Here we review landmark methods that are closely related to our work.


\subsection{High-quality $3$D Face Reconstruction}

Quality-sensitive applications of realistic $3$D avatars, such as those in the movie industry and Telepresnece in AR/VR, can recover high-quality 3D models of target individuals.
Expensive and complex multi-view capture systems~\cite{debevec2000acquiring,beeler2011high,bickel2007multi,bradley2010high,fyffe2014driving,wuu2022multiface} are used to recover high-quality geometric and appearance information.
Additionally, professional artists are employed to further clean up the recovered 3D model and improve its quality, completeness, and driveability~\cite{scanstore2023,dreamface}.
While this can achieve very high level of quality, it comes at the cost of an expensive and lengthy person-specific process.

\subsection{Parametric Face Models}

To enable accessible and cost-effective facial reconstruction, $3$D Morphable Models ($3$DMM)~\cite{blanz2023morphable} learn facial priors from a large dataset of high-quality face scans.
Such facial priors are low-dimensional parametric models for facial geometry and appearance~\cite{blanz2023morphable,booth20163d,kemelmacher2013internet,ploumpis2020towards}.
Light-weight data capture such as a monocular camera or few-shot images are then used to supervise solving an optimization problem in the low-dimensional parameter space.
However, the reduced user friction and data capture cost come at the expense of two axes -- quality and completeness.
First, the low-dimensional parameter space cannot represent identity-specific cues such as wrinkles and high-frequency appearance details,
which are crucial for a photo-realistic representation of one's identity.
Second, learning a mesh-based facial prior is limited to representing regions that are well explained by a shared topology and simple deformation models.
Therefore, such priors mainly represent the facial skin region, while missing out regions like the mouth interior, eyes and hair.




\begin{figure*}[t]
    \centering
    \includegraphics[width=\textwidth]{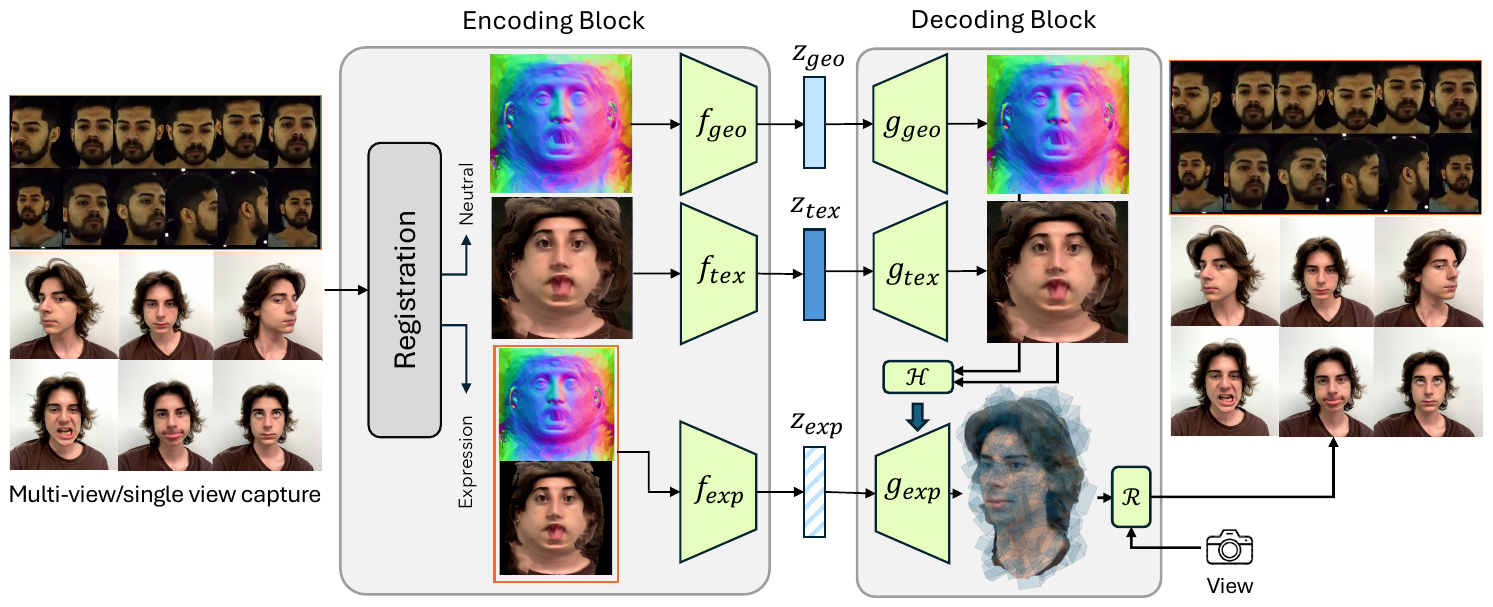}  

    \caption{Main CAAE Framework for learning the latent space for geometry and texture of avatars.}
    \label{fig:method:framework}
\end{figure*}

\subsection{Neural Rendering}
Neural Rendering~\cite{tewari2020state,tewari2022advances} techniques improved completeness and achieved photo-realistic quality by optimizing a neural representation and/or a rendering network to minimize the loss between renders and captured data.
Implicit volumetric neural representations~\cite{park2021nerfies,park2021hypernerf} can recover and render highly photo-realistic head avatars, including difficult regions such as eyes and hair, but they mainly learn a static non-animatable representation.
%
To recover dynamic and driveable avatars, a neural representation is learned on top of a parametric face model~\cite{thies2019deferred,gafni2021dynamic,kirschstein2023nersemble,zielonka2023instant}, to allow expression transfer in the parametric expression space of the template face mesh.

In contrast to using parametric models to drive an avatar, another line of work~\cite{lombardi2018deep, ma2021pixel, lombardi2021mixture, saito2023relightable} learns a high dimensional latent expression space, jointly with a latent shape and appearance code,
and train an expression encoder to map tracked expressions to the expression latent space for driveability.
Such models achieve ultra-realistic rendering under very challenging expressions, and are able to render challenging regions like eyes, teeth, tongue, mouth interior, and hair,
but they learn subject-specific models and require high-quality multi-view data~\cite{wuu2022multiface}.
More recently, Cao~\emph{et al.}~\cite{cao2022authentic} generalized codec approaches to new subjects by learning an identity-conditioned Universal Prior Model (UPM) from high-quality captures, which can be fine-tuned for phone-scans of new subjects.
However, the learned prior encodes identity information as high-dimensional multi-resolution feature maps and is not a generative model, which both limits editability and requires intricate fine-tuning strategy for personalization. 
In contrast, we propose a text-to-avatar generative model that still leverages the high quality, completeness, and driveability of Codec decoders~\cite{ma2021pixel, cao2022authentic} as a rendering framework.

\subsection{Generative Face Models}

Generating photo-realistic fictional avatars is widely desired for many applications, such as gaming and virtual AI agents in AR/VR.
Generative models trained on facial image datasets learn a strong prior and can dream up photo-realistic faces of non-existing people in 2D~\cite{karras2017progressive,karras2019style,karras2020analyzing,karras2020training}
and in 3D~\cite{pigan, eg3d, an2023panohead}, and some works showed limited driveability of generated avatars using 3DMMs~\cite{tewari2020stylerig,wu2022anifacegan,cgof,cgof++}. However, in contrast to our approach, they can only render limited view angles and cannot transfer challenging expressions, for exampling showing tongue and mouth interior.

\subsection{Text-to-3D Generation}
Recent breakthroughs in visual-language models~\cite{radford2021learning} and diffusion models~\cite{ddpm,song2020denoising,dhariwal2021diffusion} have significantly improved text-to-image generation~\cite{ramesh2022hierarchical,saharia2022photorealistic,rombach2022high,balaji2022ediffi}.
More recently, several works extend text-to-image models to generate 3D objects or scenes from text prompts by leveraging pre-trained text-to-image diffusion and optimizing 3D neural representations
that minimizes the CLIP scores between multi-view 2D renderings and text prompts~\cite{xu2023dream3d,hong2022avatarclip,michel2022text2mesh,sanghi2022clip,jain2022zero} or that employs a score distillation sampling (SDS) strategy~\cite{poole2022dreamfusion,lin2023magic3d,chen2023fantasia3d,huang2023dreamtime}.


Described3D~\cite{describe3d} is a concurrent work that generates driveable avatars from an input text prompt.
However, in contrast to our proposal, their generated avatars are limited in terms of quality and completeness.
For example, their method cannot generate challenging regions like the mouth interior, tongue, hair, head-wear or facial hair. Manual processing is required to add extra assets like hair or accessories. 
Additionally, Described3D is trained on synthetic data, so their results are far from photo-realistic. 
On the other hand, DreamFace~\cite{dreamface} can produce much more appealing driveable avatars, but their method is not a generative model.
Instead, they take a compositional approach that relies on a nearest neighbor selection of a best matching geometry from a pool of acquired assets
The selected geometry is refined through an optimization process to align the avatar with the input text prompt using pre-trained Language-Vision models.
And finally, a prompt-based hair selection is applied from a pool of 16 artist-created hair assets.
In contrast, our method relies on a data-driven prior that learns the models photo-realistic 3D head avatars. Therefore we can generate new avatars through a simple forward pass in our model. The generated avatars are photo-realistic, driveable and achieve higher completeness compared to existing methods.

\section{Methods}
\label{sec:methods}
Given a text prompt describing $\prompt$ the facial attributes of a random identity, \genca generates a photo-realistic and animatable 3D avatar that matches the text prompt.
To achieve this, we propose a two-stage framework for \genca.

In the first stage (\Cref{fig:method:framework}), we introduce a Codec Avatar Auto-Encoder (CAAE) framework.
The CAAE uses an \emph{Encoding Block}, $\mathcal{E}$, to map input images into a factorized latent space for identity and expression.
The identity latent space further broken into two latent spaces ($\zgeo, \ztex$) for the neutral geometry and texture of any given identity.
Then a \emph{Decoding Block}, $\mathcal{D}$, transforms these latent codes back into realistic images.

In the second stage (\Cref{fig:idgeneration}), we train an \emph{Identity Generation Model} to learn a mapping from an input text prompt, describing an identity, to its corresponding identity latent codes ($\zgeo, \ztex$).

During inference, given a text description of facial attributes, \genca utilizes the {Identity Generation Model} to sample an identity latent code and employs the Decoding Block $\decblock$ to convert both the identity and expression codes into a photo-realistic drivable 3D avatar.

\subsection{\label{subsec:caae}Codec Avatar Auto-Encoder (CAAE)}
\subsubsection{\label{subsubsec:preliminaries}Preliminaries}

Our CAAE extends the Universal Prior Model (UPM) for 3D faces proposed by~\cite{cao2022authentic} (\Cref{fig:method:framework}). 
%
The UPM is composed of an expression encoder $\Eexp$, an identity-conditioned expression decoder $\Dexp$, and a hyper-network $\hypernet$.
The expression encoder $\Eexp$ takes as input per-frame geometry and texture ($\Gexp, \Texp)$ and generates a universal expression code $\zexp$ that is shared across identities. 
The hyper-network $\hypernet$ extracts identity features from the average neutral geometry and texture ($\Gneu, \Tneu$) and modulates the weights of the expression decoder $\Dexp$.
The expression decoder $\Dexp$ then decodes the expression code $\zexp$ into neural volumetric primitives~\cite{lombardi2021mixture} that are rendered from any camera pose $\campose$ into photo-realistic images $\Irecon$ using a renderer $\renderer$.
\begin{align}
    \Irecon = \renderer(\Dexp(\zexp | \hypernet(\Tneu, \Gneu)), \mathcal{C}), \quad
    \zexp = \Eexp (\Texp, \Gexp)
    \label{equation:render}
\end{align}
%
%


While~\cite{cao2022authentic} train the UPM solely on a few hundreds of high-quality multi-view dome captures, we opt to re-train the UPM by including an additional large-scale dataset of phone captures that follow the phone scan process proposed in~\cite{cao2022authentic}.
To bridge the domain gap between the high-quality multi-view dome captures and the simple phone captures, we apply a discriminator on the expression codes $\zexp$ to encourage a unified latent space for both datasets.
This additional data injection improves the UPM's generalization ability to reconstruct diverse identities.
Both the phone and multi-view datasets are registered using the method proposed in~\cite{cao2022authentic}.

We denote that the UPM from~\cite{cao2022authentic} is an auto-encoder framework and is not a generative model, as it neither learns avatar appearance/identity distribution, nor supports new avatar sampling.
In contrast, the first stage of our \genca (\Cref{fig:method:framework}) extends the UPM by introducing encoders and decoders for the average neutral geometry and texture ($\Gneu, \Tneu$), forming the Encoding and Decoding Block of our CAAE (Sec.~\ref{subsec:caae}).
And the second stage of \genca (Section~\ref{sec:igm}) trains a generative model for the identity latent space.


\begin{figure}[t]
\centering
  \includegraphics[width=0.5\textwidth]{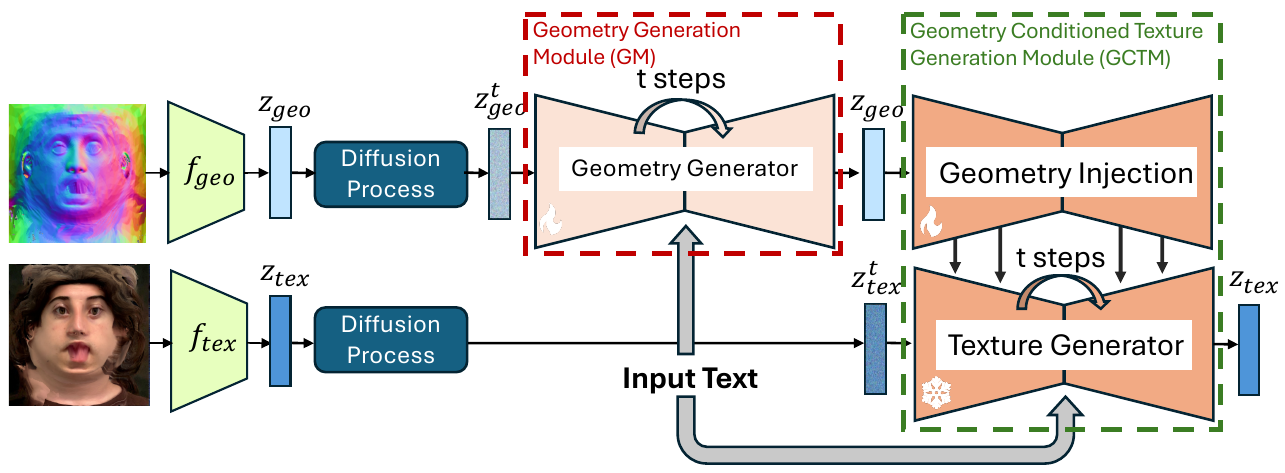}
  \caption{\textbf{Training Pipeline of the Identity Generation Model}, Geometry generator Module (GM): Generates $z_{geo}$ of realistic geometries based on text descriptions. Geometry Conditioned Texture Generation (GCTM): Generates $z_{tex}$ of high quality texture, consistent with conditioned geometry, based on the text descriptions.}
  \label{fig:idgeneration}
\end{figure}
\subsubsection{\label{sec:encoding}Encoding Block}

As shown \Cref{fig:method:framework}, the Encoding Block $\mathcal{E}$ is composed of a Registration Module $\registration$, an encoder for neutral geometry UV map $f_{geo}$, an encoder for neutral texture UV map $f_{tex}$, and the expression encoder $\Eexp$ (introduced in~\Cref{subsubsec:preliminaries}). 

A single or multiple input images $I_{inp}$ of a subject is categorized into either neutral expression and expressive expression segments based on the capture script. 
The Registration Module $\registration$ then reconstructs the per-frame geometry and the associated unwrapped texture as follows: 
\begin{align}
    \Tneu, \Gneu, \Texp, \Gexp = \registration(I_{inp})
    \label{equation:registration}
\end{align}
Specifically, it computes the average neutral geometry map $\mathcal{G}_{neu}$, neutral texture map $\mathcal{T}_{neu}$ using the captured segment with neutral expression.
Using the capture segment with expressive expressions,
the per-frame expression geometry and texture maps $\Gexp$ and $\Texp$, as well as the camera view $\mathcal{C}$ for each frame.

After the registration step, the Encoding Block $\mathcal{E}$ encodes $\mathcal{G}_{neu}$, $\mathcal{T}_{neu}$, $\Gexp$, $\Texp$ into the latent space:
\begin{align}
    \zgeo, \ztex, \zexp &= \mathcal{E}(\Tneu, \Gneu, \Texp, \Gexp)
\end{align}
where $\zgeo$, $\ztex$, $\zexp$ are the latent codes for geometry, texture and expression respectively.
In particular, the identity geometry latent code $\zgeo$ and the identity texture latent code $\ztex$ are computed by:
\begin{equation}
    \zgeo = \Egeo(\Gneu), \qquad \ztex = \Etex(\Tneu)    
    \label{equation:enc_neu}
\end{equation}
and the expression latent code $\zexp$ is obtained as described in~\Cref{equation:render}.

\subsubsection{\label{sec:decoding} Decoding Block}

%
Given a registered camera pose $\mathcal{C}$, the Decoding Block $\mathcal{D}$ maps the latent codes from the Encoding Block $\mathcal{E}$ to a rendered image:
\begin{align}
     \Irecon = \mathcal{D}(\zgeo, \ztex, \zexp, \mathcal{C})
\end{align}

In particular, the Decoding Block $D$ contains a neutral texture decoder $g_{tex}$ and a neutral geometry decoder $g_{geo}$, which take the neutral geometry and texture latent codes and reconstruct the associated registered UV maps:
\begin{align}
    \label{eq:decode_id_tex}
    \hat{\mathcal{T}}_{neu} &= g_{tex} (\ztex)\\
    \label{eq:decode_id_geo}
    \hat{\mathcal{G}}_{neu} &= g_{geo} (\zgeo)
\end{align}

The Decoding Block also contains an expression decoder $g_{exp}$ and a hyper-network $\mathcal{H}$ as introduced in~\Cref{subsubsec:preliminaries}. 
We employ the hyper-network $\mathcal{H}$ to extract feature maps from the reconstructed $\hat{\mathcal{T}}_{neu}$ and $\hat{\mathcal{G}}_{neu}$, which is then used to modulate $g_{exp}$. The output of $g_{exp}$ is eventually rendered into an image $\Irecon$ by the renderer $\mathcal{R}$ given a registered camera pose $\mathcal{C}$.

\begin{align}
    \Irecon = \mathcal{R}(g_{exp}(\zexp |\mathcal{H}(\hat{\mathcal{T}}_{neu}, \hat{\mathcal{G}}_{neu})), \mathcal{C})
    \label{equation:decode:render}
\end{align}



\subsubsection{Loss Functions}
To train the $\Eexp$ and $\Dexp$, we first employ the loss functions $\mathcal{L}_{upm}$ from UPM~\cite{cao2022authentic} as the reconstruction loss.
To train the identity auto-encoders, including $\Etex$, $\Dtex$, $\Egeo$ and $\Dgeo$, we further compute the $\mathcal{L}_1$ loss for the reconstructed geometry and texture:
\begin{equation}
    \mathcal{L}_{geo} = |\Gneu - \hat{\mathcal{G}}_{\scriptsize \mbox{neu}}|, \qquad
    \mathcal{L}_{tex} = |\Tneu - \hat{\mathcal{T}}_{\scriptsize \mbox{neu}}|
\end{equation}
To regularize the latent space as a normal distribution, we also minimize the Kullback–Leibler (KL) divergence $\mathcal{L}_{KL}$ between the learned neutral geometry and texture latent distribution and a standard Gaussian distribution.

\subsection{Identity Generation Model}
\label{sec:igm}

The Codec Avatar Auto-Encoder (CAAE) introduced in~\Cref{subsec:caae} maps facial images into a smooth latent space, and reconstructs the latent code into images.
Following Latent Diffusion Models \cite{rombach2022high}, we train a diffusion model in the identity latent space $z_{id} = \langle\ztex, \zgeo\rangle$. Dubbed as the Identity Generation Model, this diffusion model maps a noise map to a latent identity code conditioned on a text prompt $\prompt$, which can be decoded by $\mathcal{D}$ and rendered into high-fidelity photo-realistic images.

The Identity Generation Model comprises two primary components: the Geometry Generation Module (GM) and the Geometry Conditioned Texture Generation Module (GCTM). The schematic of the Identity Generation Model is shown in Fig.~\ref{fig:idgeneration}.


\begin{figure*}[h]
    \centering
    \includegraphics[trim = 0 0 0 0,  width=0.9\textwidth]{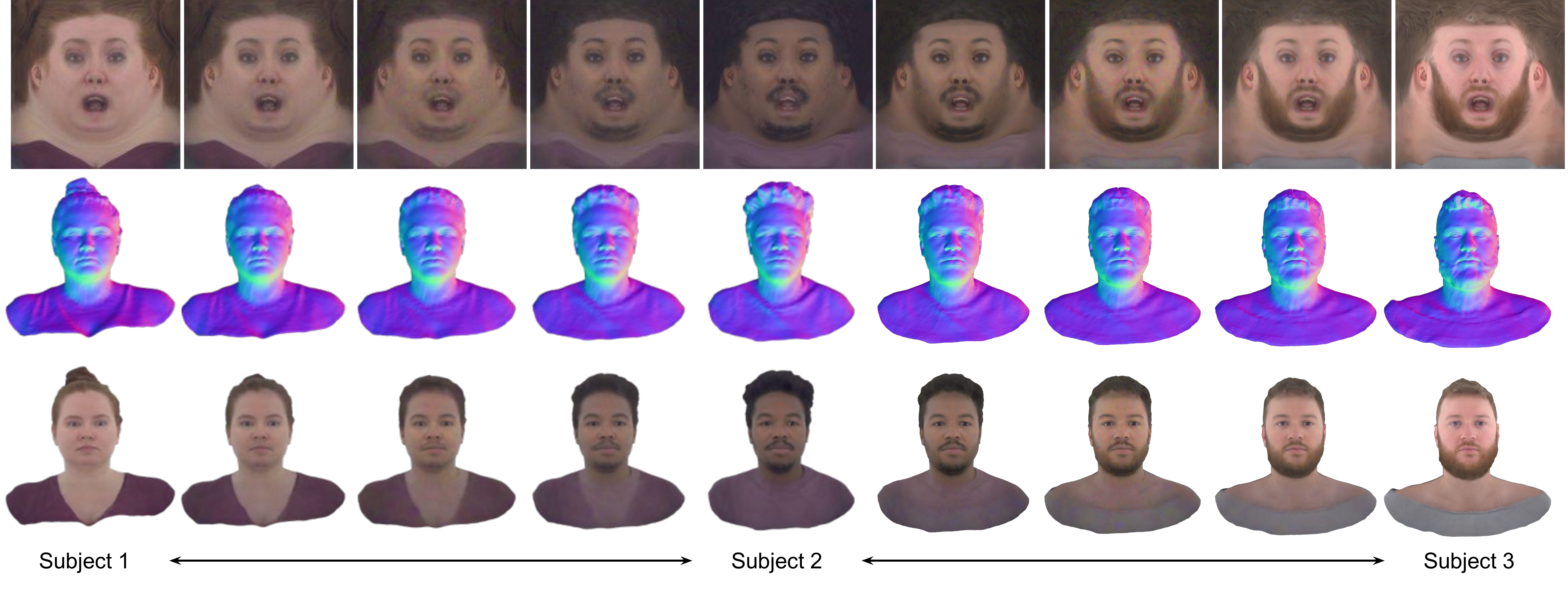}
    \vspace{-7mm}
    \caption{Smooth linear interpolation among the geometry and texture latent codes.}
    \label{fig:interpolation}
\end{figure*}

\subsubsection{Geometry Generation Module (GM)}
Given a text prompt $\prompt$ and a noise map $\zgeo^t$, GM $\epsilon_\theta$ aims to denoise from $\zgeo^t$ to $\hat{z}_{geo}^{t-1}$. By repeating this process recursively, GM eventually yeilds $\hat{z}_{geo}^{0} \sim f_{geo}(\zgeo|\mathcal{G}_{neu})$, which can then be decoded by~\Cref{eq:decode_id_geo}.

For training this latent diffusion model, we follow~\cite{ddpm,rombach2022high} to perform the diffusion process manually to construct supervision. Specifically, given a latent code $\mathbf{z}_{geo}$, we first add noise to the $t$-th time step:
\begin{align}
    \mathbf{z}_{geo}^t = \sqrt{\bar{\alpha}_t}\mathbf{z}_{geo} + \sqrt{1-\bar{\alpha}_t}\mathbf{\epsilon},
\end{align}
The network $\epsilon_\theta$ then takes the noised latent code $\mathbf{z}_t$ and time step $t$ to predict the noise
\begin{align}
    \hat{\mathbf{\epsilon}}_{geo}^t = \epsilon_\theta(\mathbf{z}_{geo}^t, t),
    \label{eq:unet} 
\end{align}
Finally,  we re-sample the latent code $\mathbf{z}_{geo}^{t}$ with the predicted noise $\hat{\mathbf{\epsilon}}_t$
\begin{align}
    \hat{\mathbf{z}}_{geo}^{t-1} = h_\eta(\mathbf{z}_{geo}^t, \hat{\mathbf{\epsilon}}_{geo}^t). \label{eq:resample} 
\end{align}

The \Cref{eq:resample} and \Cref{eq:unet} are recursively solved to get an estimate $\hat{\mathbf{z}}_{geo}^0$ which can be used to produce the estimated geometry by using the VAE decoder via~\Cref{eq:decode_id_geo}

\subsubsection{Geometry Conditioned Texture Generation Module (GCTM)}
Directly applying the structure of GM for texture generation results in poor convergence and severe semantic misalignment, as shown in~\Cref{fig:exp:ablation}.
Inspired by ControlNet ~\cite{controlnet}, we devise a Geometry Conditioned Texture Generation Module (GCTM), where the Texture Generator $\epsilon_\phi$ takes the geometry information from the Geometry Injection $\epsilon_\psi$ module, and generates a corresponding texture latent code.

Specifically, given a geometry latent code $\zgeo$, we extract feature maps with the Geometry Injection module $\epsilon_\psi$ and inject the features to the Texture Generator $\epsilon_\phi$, to predict the noise in $\ztex^t$:

\begin{align}
    \hat{\epsilon}_{tex}^t = \epsilon_{\phi}(\ztex^t, t|\epsilon_{\psi}(\zgeo))
\end{align}

$\hat{\epsilon}_{tex}^t$ is then used for denoising, and by repeating this process, $\epsilon_{\phi}$ eventually generates $\hat{z}_{tex}^{0}$, which can be decoded via~\Cref{eq:decode_id_tex}.

We observed that training with GCTM results in better topological alignment between texture and geometry compared to standalone texture generation.

\subsubsection{Loss Functions} We follow~\cite{ddpm} and use the latent diffusion loss for both the GM and GCTM modules:

\begin{equation}
\mathcal{L}_{ldm} := \mathbb{E}_{\mathcal{E}(I_{inp}), \epsilon \sim \mathcal{N}(0, 1),  t}\Big[ \Vert \epsilon^t - \hat{\epsilon}^t \Vert_{2}^{2}\Big] \, .
\label{eq:ldmloss}
\end{equation}

\subsection{Inference}

After training, to generate a new avatar, we first use GM to generate a high-quality UV position map based on the text description. This map is then used as an extra condition, alongside the text prompt, to generate the corresponding texture via GCTM.
Their ouputs, along with a selected expression code $\zexp$, are decoded by the Decoding Block and rendered into the final avatar images.


\section{Experiments}
\label{sec:exp}

\subsection{Data}
\label{subsec:data}
For training \genca we use images data from two different sources. The first source of data is from a capture dome setup where synchronized multi-view cameras are setup and we capture an extensive set of subjects expressions as the subject follows a pre-defined expression script. 
The second source of data is the $12000$ phone scan data which are acquired by a single view tripod video capture using iPhone $13$ of the frontal face rotating in $45$ degree span while maintaining the neutral expression. Both the multi-view captures as well as single view captures are collected with mindset to covering a diverse population of identities.

Both the data sources have associated attributes, and we use a large language model API for generating descriptions, more details in Supplementary.


\subsection{Implementation Details}

Our implementation is built upon Latent Diffusion Model (LDM). For both the GM and GCTM, we initialize the VAE with the weights pre-trained on in-the-wild image datasets. The AdamW optimizer with a fixed learning rate of $1\times {10}^{-5}$ is employed, and the generation for geometry and texture requires $40$ iterations in total. It takes about $40$ seconds to generate a face on a single NVIDIA A100 GPU. We refer readers to the supplementary material for more details.

\begin{figure*}[t]
    \centering
    \includegraphics[trim = 0 0 0 0,  width=\textwidth]{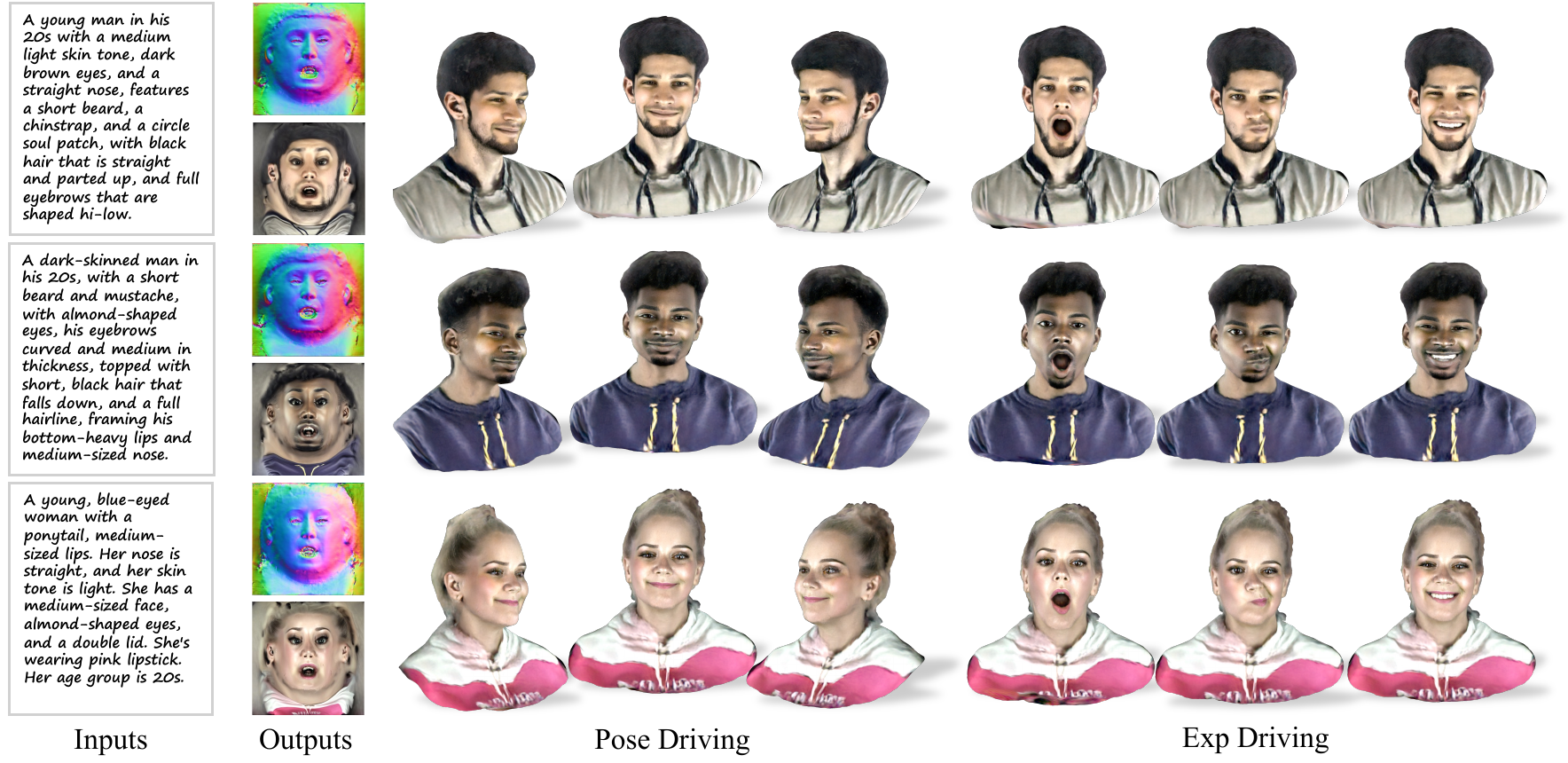} 
    \caption{\textbf{Generation Results}: Qualitative results generated from the captions provided in leftmost column.}
    \label{fig:exp:gen}
\end{figure*}
\begin{table*}
\centering
\captionsetup{justification=centering} 
\caption{Comparison with three state-of-the-art Methods. The three numbers for User Study denote: Semantic Alignment/Visual Appealing/Overall Preference}
\vspace{-3mm}
\begin{tabular}{lccccc}
\toprule
Method                      & CLIP Score & Aesthetic Score & HPS v2 & User Study \\
\midrule
Describe3D~\cite{describe3d}      & 17.71      & 4.581          & 0.1510 &        0.61\% / \underline{7.88\%} / 4.85\%    \\
SOTA-M1      & 17.70 & \underline{4.720} & 0.1605  & 2.42\% / 7.27\% / 6.06\%          \\
SOTA-M2          & \textbf{17.83}      & 4.471          & \underline{0.1635} & \underline{32.73\%} / 1.21\% / \underline{6.06\%} \\
Ours                     & \underline{17.82}      & \textbf{4.799}          & \textbf{0.1731} & \textbf{64.24\%} / \textbf{83.64\%} / \textbf{83.03\%} \\
\bottomrule
\end{tabular}
\vspace{-3mm}
\label{tab:sotacomparison}
\end{table*}


\begin{table}
\centering
\captionsetup{justification=centering} 
\caption{
Ablation study of the proposed method.
}
\vspace{-3mm}
\resizebox{0.5\textwidth}{!}{%
\begin{tabular}{lccccc}
\toprule
 & Condition & NeuRender & CLIP Score & Aes. Score & HPS v2  \\
\midrule
1 & None & Y &  17.02    & 4.462  &  0.1073 \\
2 & Disp & Y &  17.45    & 4.737  &  0.1675  \\
3 & Norm & Y &  \textbf{17.82 } &  \textbf{4.799}   &  \textbf{0.1731}   \\
4 & Norm & N & 16.78  & 4.682  & 0.1530\\
\bottomrule
\end{tabular}
}
\vspace{-5mm}
\label{tab:AblationStudy}
\end{table}


\subsection{Generation Results}
\subsubsection{Interpolation}
In Fig.~\ref{fig:interpolation},
we show interpolation results to demonstrate the latent space of the trained CAAE can be interpolated smoothly, which lays a solid foundation for the Identity Generation Model.

\subsubsection{Text conditioning generation}
As shown in Fig.~\ref{fig:exp:gen}, \genca takes a sentence in input and generate corresponding Avatars, which is faithful to the input description and can be driven to perform various wild expressions. \genca has the capability to produce avatars with a wide range of diversity, encompassing aspects such as ethnicity, age, and appearance.

\subsubsection{Additional Applications}
Please note that \genca is versatile and can be applied to various downstream tasks, such as single or multi-image registration and text-based 3D avatar editing (e.g., adjusting skin tone, goatee, clothing, hair color, etc.). Due to space constraints, the results are included in the supplementary material. We encourage you to refer to it for detailed outcomes.

\begin{figure*}[h]
    \centering
    \includegraphics[trim = 0 0 0mm 0,  width=\textwidth]{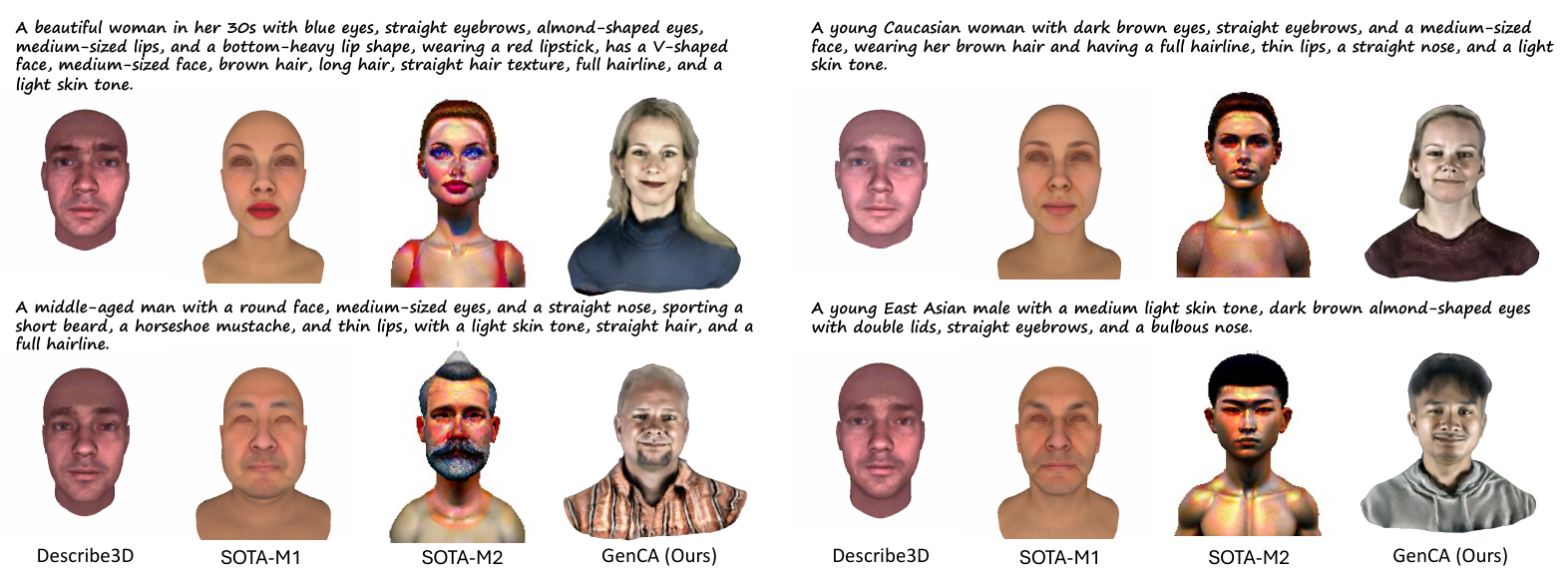}
    \vspace{-0.5cm}
    \caption{Qualitative comparison of \genca generation with three state-of-the-art methods. Compared to other methods, \genca produces more comprehensive and photorealistic avatars given the same text descriptions.}
    \label{fig:exp:cmp_sota}
\end{figure*}

\subsection{Evaluation Metrics}
\label{sec:evalmetrics}
We evaluate \genca with several quantitative evaluation metrics common to generative models. These metrics include the contrastive language–image pretraining (CLIP) score \cite{hessel2021clipscore}, Aesthetic Score \cite{murray2012ava} and human preference score (HPS) \cite{wu2023hps}. The CLIP score provides semantic accuracy of generation given a text caption, the Aesthetic Score evaluates the aesthetic quality of generated images and the HPS demonstrates the alignment between text-to-image generation and human preferences. In addition to these evaluation metrics we also conduct an unbiased user study.

\subsection{Comparison with State-of-the-Art Methods}
We compare with three state-of-the-art methods Describe3D\cite{describe3d}, and other two SOTA methods M1 and M2 that uses generative approaches to create text-driven avatars. The qualitative comparison in Fig.~\ref{fig:exp:cmp_sota}, shows superiority of \genca in generating photo-realistic avatars and following the text description accurately. The quantitative comparisons are shown in ~\Cref{tab:sotacomparison}, \genca performs similarly with SOTA-M2 on the CLIP score but outperforms all for the other three metrics, especially on the User Study result, \genca achieves the best scores by a significant margin.

\subsection{Ablation Study}
\begin{figure}
\centering
\includegraphics[width=0.44\textwidth]{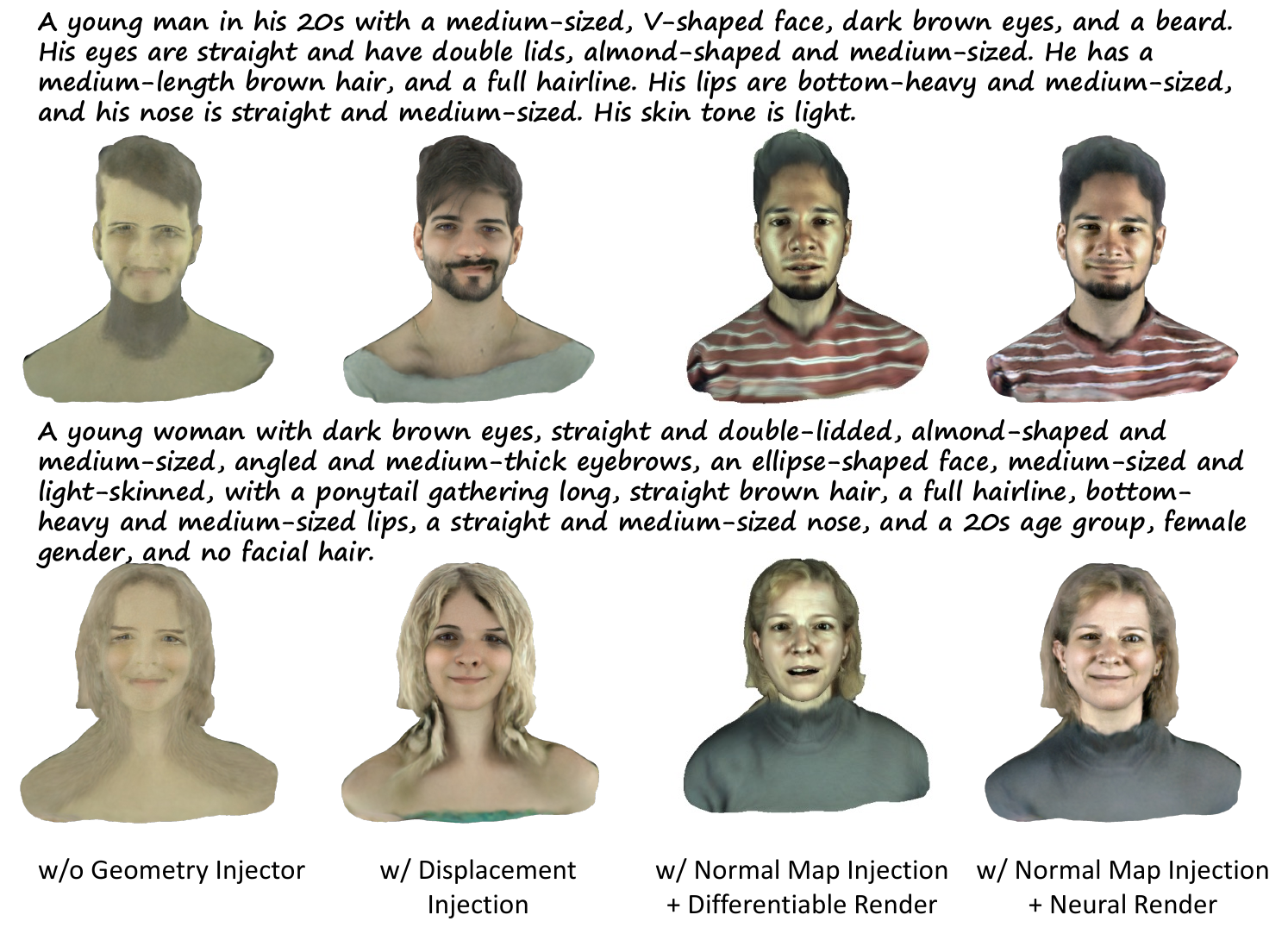}
\vspace{-0.5cm}
\caption{Ablation study by disabling different parts of the proposed method. }
\vspace{-0.35cm}
    \label{fig:exp:ablation}
\end{figure}

We perform ablation studies of different components of the proposed method as shown in Fig.~\ref{fig:exp:ablation} and~\Cref{tab:AblationStudy}. For quantitative ablation, we reuse the metrics introduced in~\Cref{sec:evalmetrics}.
We mainly explore the effects of different geometry conditioning (GeoCond) in the texture generation and the effects of the neural renderer (NeuRender).
Specifically, the model without any geometry conditioning (None), as shown in the first row of~\Cref{tab:AblationStudy} and the first column of Fig.~\ref{fig:exp:ablation}, fails to learn a plausible texture, with blurry artefacts, leading to an extremely low human preference score.
Taking displacement map (Disp) for geometry injection, as shown in the second row of~\Cref{tab:AblationStudy} leads to plausible texture and significant improvement in all the metrics. However, as shown in the second column of Fig.~\ref{fig:exp:ablation}, the texture is not well-aligned with the geometry, and has severe distortion in the face. In our final design (Norm), we propose to compute the normal map as the geometry conditioning, which leads to the best performance both qualitatively and quantitatively.

Furthermore, we evaluate the effectiveness of our neural renderer with another comparison in row 3 and 4  of~\Cref{tab:AblationStudy} and column 3 and 4 of Fig.~\ref{fig:exp:ablation}. Even with the same geometry and texture map generated by our GenCA, using a traditional graphics-based renderer~\cite{paszke2017automatic} leads to unrealistic artifacts in the skin, whereas the proposed neural rending block yields visually appealing effects.

\section{\label{sec:conclusion}Conclusion}

We propose GenCA, a text-guided generative model capable of producing photorealistic facial avatars with diverse identities and comprehensive features. With
complete details, such as hair, eyes, and mouth interior, which can be driven
through. We also showcase a variety of downstream applications enabled by GenCA, including avatar reconstruction from a single image, editing and inpainting. Finally, we achieve superior performance/quality in comparison to other state-of-the-art methods.



\appendix
\medskip
\medskip
\noindent{\LARGE\bf Appendices}

\section{Inference Architecture}

Once trained, GenCA generates neutral texture and neutral geometry latent codes using input text prompt and random noise. Once we get the $\zgeo$ and $\ztex$, we pass it through the decoding block $\decblock$ to obtain the UV maps $\hat{\mathcal{T}}_{neu}, \hat{\mathcal{G}}_{neu}$, which is then rendered using the expression and view parametrization:

\begin{align}
    \Irecon = \mathcal{R}(g_{exp}(\zexp |\mathcal{H}(\hat{\mathcal{T}}_{neu}, \hat{\mathcal{G}}_{neu})), \mathcal{C})
\end{align}

\begin{figure*}[h!]
    \centering
    \includegraphics[trim = 0 0 0 0,  width=0.95\textwidth]{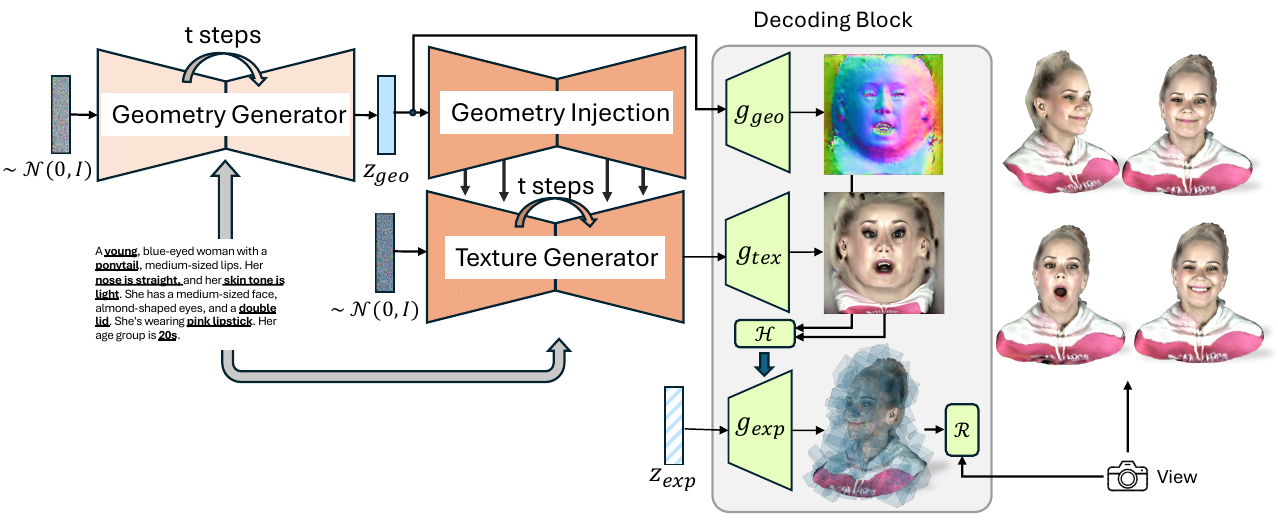}
    \caption{Inference for the GenCA. We pass in random noise and a text prompt to generate the geometry and texture neutral codes. This is then passed through the decoding block or obtaining view and expression parametrized renderings of the generated avatar.}
    \label{fig:sup:inference}
\end{figure*}

The inference process is shown in \Cref{fig:sup:inference}

\section{Additional Implementation Details}

In training the Geometry Generator, we set the resolution to $1024\times1024$, and set batch size to $12$. We trained the Geometry Generator on use $8$ NVIDIA A100 GPUs for $8$ hours.

When training the Geometry-Conditioned Texture Generator, we take the ground-truth geometry map as the input condition, and the corresponding texture map as the supervision.
The resolution of the Texture Generator is also set to $1024\times1024$, whereas the batch size is $4$, and it takes $12$ hours to train on $8$ NVIDIA A100 GPUs.

\section{Dataset}

\begin{figure*}[!h]
		\centering
  \vspace{-7mm}
	\includegraphics[width=\textwidth]
 {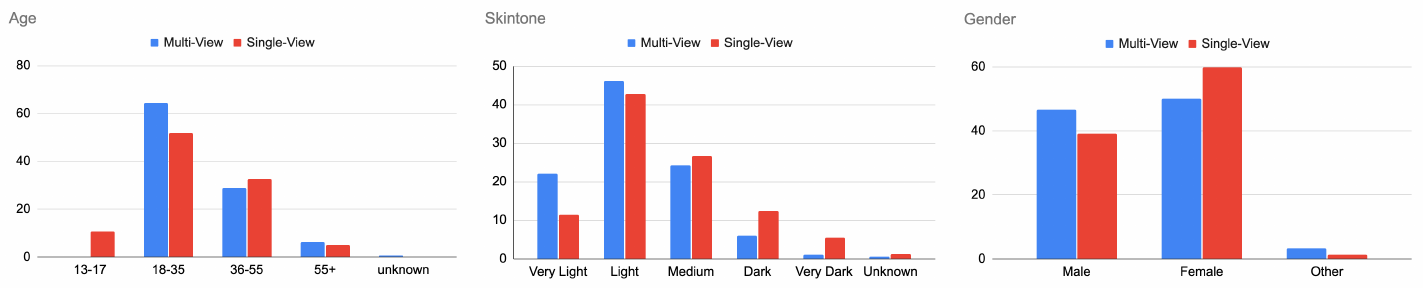}
\vspace{-0.5cm}
\caption{Data percentages with respect to three different attributes, age, skintone and gender in the multi-view and single-view captures used for \genca training.}
    \label{fig:data:attributes}
\end{figure*}

For each subject, we use a Visual Language Model (VLM) to annotate the global attributes like \textit{age and gender}, and the local attributes for the \textit{eyebrows, eyes, glasses, lips, skin tone, hair, nose, face shape, face size, facial hair, eyebrow style, eyebrow thickness, eye color, eye direction, eye lid type, eye shape, eye size, glasses frame, glasses size, glasses style, lip shape, lip size, lip stick color etc}. As an example, Fig.~\ref{fig:data:attributes} shows the percentage distribution of both the muti-view dome dataset and the single-view phone capture with respect to three attributes (age, gender, and skintone).
Eventually, we combine a Large Language Model (LLM) to summarize all these information into a sentence to describe the subject.

\begin{figure}[t]
		\centering
  \vspace{-7mm}
	\includegraphics[width=9cm]
 {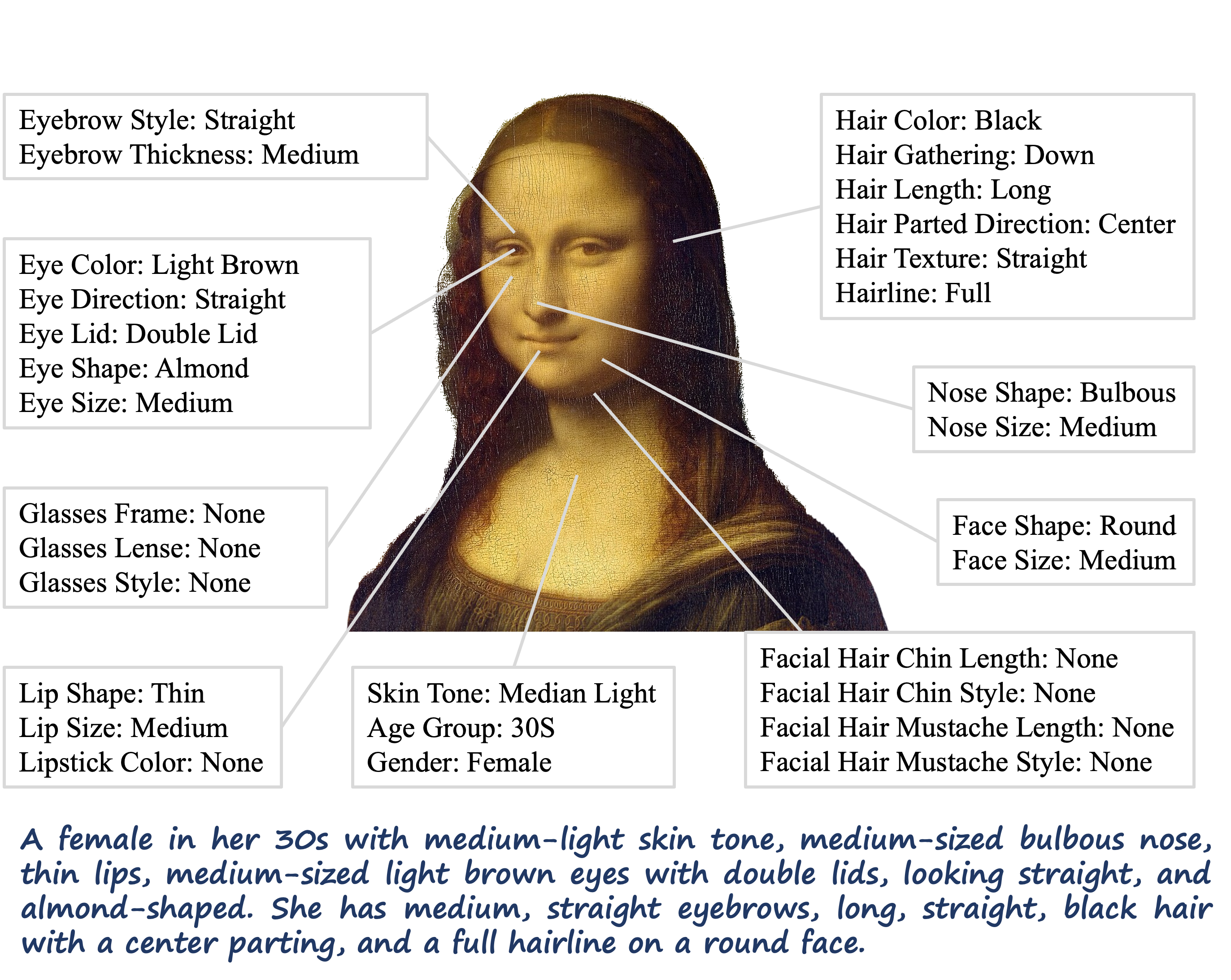}
\vspace{-0.5cm}
\caption{Example of extracting attributes and generating text caption.}
    \label{fig:data:anno}
\end{figure}

\begin{figure*}[h!]
    \centering
    \includegraphics[trim = 5mm 5mm 0mm 5mm, width=0.99\textwidth]{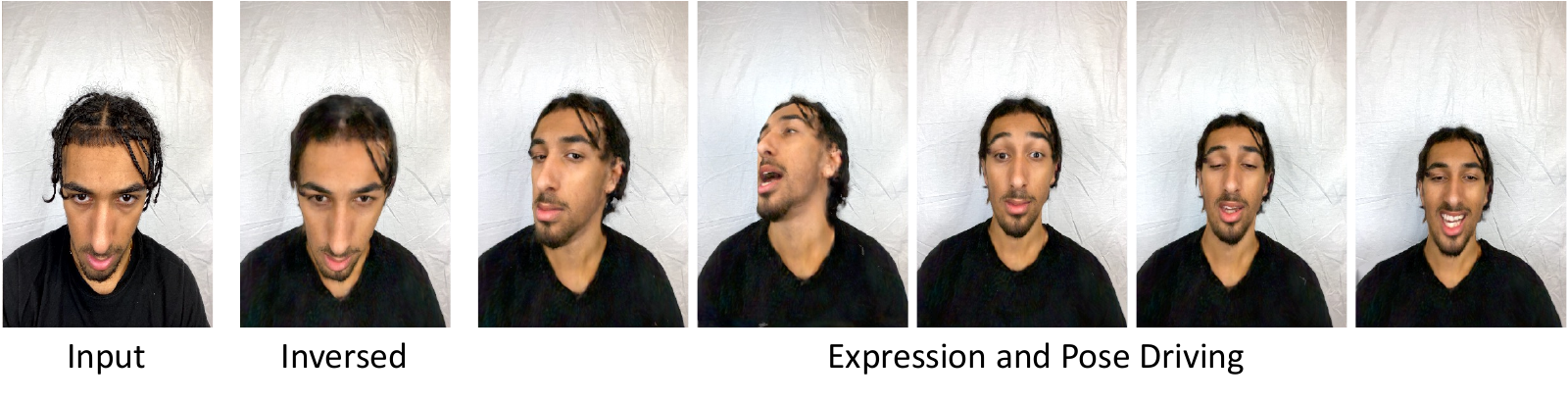}
    \caption{Given the input image in the first column, we perform inversion to reconstruct the full Codec Avatar (second column), which can be driven with different expressions and rendered from different poses.}
    \label{fig:exp:inversion}
\end{figure*}

\begin{figure*}[h!]
    \centering
    \includegraphics[trim = 0 0 0 0,  width=0.95\textwidth]{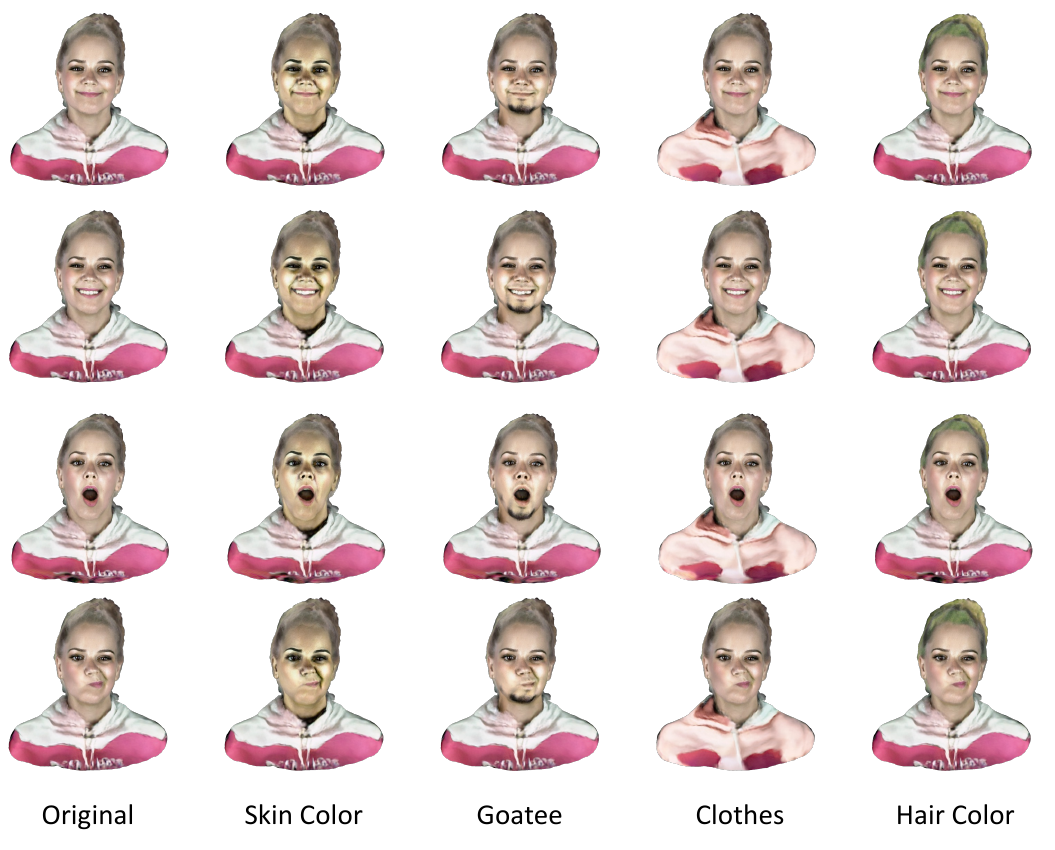}
    \caption{Editing results of generated avatar in the first column with different expressions.}
    \label{fig:exp:editing}
\end{figure*}

\section{Additional Results}
The Identity Generation Model with its rich identity latent space allows us to leverage the power of generative model for several applications including personalization and editing. This in turn allows for a flexible and quick way to control the avatar's appearance while still maintaining the expression driving and generalization via the UPM. 

\subsection{Single/Multi-Image Personalization} Given, we have a latent space for identities we can invert into that space via the decoding block and find the $z_{geo}$ and $z_{tex}$ corresponding to a provided single/multi image input. More specifically, we use a single or multi-image (Fig.~\ref{fig:exp:inversion}) to get an approximate UV texture and geometry, which allow us to get an initial estimate of the latent codes using the encoding block. We then leverage UPM losses (\cite{cao2022authentic}) to supervise on the input data and backpropogate into the identity latent space, similar methods are used for GAN inversion \cite{abdal2019image2stylegan}.

\subsection{Editing} Given any generated or personalized avatar, we can perform global and local editing. For global appearance editing, we solely utilize text conditioning with GCTM, which alters the texture while maintaining the geometry. For local editing, we use semantic masks in the UV space to perform in-painting \cite{lugmayr2022repaint}. Specifically, given a mask $M$, we perform the following edit on the UV space $F_{out} = (1-M) \times F_{edit} + M\times F_{gen}$, here $F_{edit}$ and $F_{gen}$ are the edited and generated geometry or texture feature respectively. This type of local editing enables us to selectively modify the avatar's geometry (e.g., hairstyle) and texture (e.g., hair color) attributes, while leaving the rest of the avatar unchanged, as shown in Fig.~\ref{fig:exp:editing}

\section{Limitation}
Currently, \genca generates texture with baked in lighting information, making it challenging to relight the generated avatars with new lighting conditions.
Additionally, our model still struggles with generating fine-grained details for regions such as hair or sharp details for the clothing regions.
Finally, as our model is based on the Codec Avatar \citet{cao2022authentic}, it inherits some of its limitations, including the inability to model complex accessories like glasses or intricate and longer hairstyles.

\section{Ethical Concerns and Social Impact}

\genca introduces a powerful method for generating plausible and photo-realistic 3D avatars based on text prompts, with the ability to control both view and expression. While \genca has proven to be beneficial for many downstream tasks, it also raises significant ethical concerns, particularly regarding potential misuse.

One major risk is the possibility of personification, where the generated avatars could be used to impersonate real individuals in a misleading or harmful way. Although our method does not offer indistinguishable, pixel-perfect 2D video synthesis and does not model several critical parameters such as lighting and background—key elements typically exploited in malicious impersonation—these limitations do not entirely eliminate the risk.

To mitigate these concerns, we have implemented strict ethical guidelines in the development and deployment of \genca. The model has been trained exclusively on data collected with prior approval, ensuring that only consenting subjects were involved in our research. Moreover, our method is intended solely for legitimate uses, such as in creative industries and virtual communication, and we strongly discourage any application that could lead to ethical violations or harm to individuals.

Equally important is the advancement of methods to detect fake content. Recent works such as \cite{wang2020cnn, ojha2023towards, cozzolino2024raising} have addressed the challenge of distinguishing real from fake images. Notably, the work by Davide et al.\cite{cozzolino2024raising} demonstrates that a CLIP-based detector, trained on only a handful of example images from a single generative model, exhibits impressive generalization abilities and robustness across different architectures, including recent commercial tools such as DALL-E 3, Midjourney v5, and Firefly.

As generative models continue to improve and access to these models becomes increasingly widespread, ongoing innovation in detection methods will be essential. It is crucial to continue research in the field of fake image detection to keep pace with the rapidly evolving domain of image synthesis. Our model, \genca, could potentially contribute to detecting fake content by providing realistic generated images and videos to enhance detection models.

{
    \small
    \bibliographystyle{ieeenat_fullname}
    \bibliography{main}
}

\end{document}